# Predicting Driver Intention Using Deep Neural Network

**Mahdi Bonyani[a] , Mina Rahmanian[b], Simindokht Jahangard\***

[a] *Department of Computer Engineering, University of Tabriz, Iran* m_bonyani96@ms.tabrizu.ac.ir

[b] *Department of Computer Engineering, Shiraz Branch, Islamic Azad University, Shiraz, Iran* Minarahmanian17@gmail.com

[*]*Department of Robotics Engineering, Amirkabir University of Technology, Iran* s_jahangard@aut.ac.ir

## Abstract

To improve driving safety and avoid car accidents, Advanced Driver Assistance Systems (ADAS) are given significant attention. Recent studies have focused on predicting driver intention as a key part of these systems. In this study, we proposed new framework in which 4 inputs are employed to anticipate diver maneuver using Brain4Cars dataset and the maneuver prediction is achieved from 5, 4, 3, 2, 1 seconds before the actual action occurs. We evaluated our framework in three scenarios: using only 1) inside view 2) outside view and 3) both inside and outside view.  We divided the dataset into training, validation and test sets, also K-fold cross validation is utilized. Compared with state-of-the-art studies, our architecture  is faster and achieved higher performance in second and third scenario. Accuracy, precision, recall and f1-score as evaluation metrics were utilized and the result of 82.41%, 82.28%, 82.42% and 82.24% for outside view and 98.90%, 98.96%, 98.90% and 98.88% for both inside and outside view were gained, respectively.

**Keywords:** Driving maneuvers prediction · Advanced Driver Assistance System (ADAS) · Deep neural networks · DenseNet · LSTM

## 1. Introduction

In recent years, World Health Organization[1] (WHO) has reported that around  1.35 million people are killed in road accidents annually worldwide. However, these statistics only include fatal injuries to passengers due to the car accidents [1]. Also, maneuvers such as changing lanes and turning while driving play an important role in road accidents [2]. To reduce the number of such fatal accidents, a mechanism that can warn drivers before performing a dangerous maneuver can be helpful [3].

Over the years, many industries and academies have focused on the development of autonomous vehicles. Google and Tesla are among the leading industries that have made significant progress in the field of autonomous vehicles [4]. Similarly, over the past few years, researchers have extensively studied semiautonomous assistive driving, and as a result, the Advanced Driver Assistance System (ADAS) promises to prevent accidents, reduce greenhouse gas emissions, transmit inability to move, and driving-related stress relief is being developed [4, 5]. These systems are designed and equipped with sensory systems to understand information about road and driving conditions, assess hazards and driving warnings, and provide visual / audio assistance to drivers [6]. Consequently, Advanced Driver Assistance Systems can increase road safety by preventing dangerous maneuvers. These systems approach achieved this goal by taking control of the vehicle or by providing additional information to the driver or identifying the driver's target by the characteristics of the driver's behavior or driving environment. It has been proven that, with the help of advanced deep learning techniques and computer vision, it is possible to predict maneuvers a few seconds in advance with high accuracy by monitoring the driver's behavior inside the car (e.g. head pose, eye movement) and using the car's own information (e.g. speed) and environment (lanes configuration, presence of intersections, etc.) [2].

---

[1] Road traffic injuries, February 7. 2020. Accessed on: Feb. 9, 2020. [Online]. Available: https://www.who.int/news-room/fact-sheets/detail/road-traffic-injuries

However, strong and flawless fully automatic vehicles have not yet been widely used in urban environments. Accidents caused by immature systems can lead to driver distrust and irreparable damage [5]. Therefore, solving unresolved and advanced challenges is essential [5]. In this paper, some of the aforementioned problems are alleviated for the sake of coming up with a relatively accurate system for recognizing driver intention via deep learning by means of Convolutional Neural Networks (CNNs). A novel architecture is proposed in which in-cabin and out-cabin view can be fed into model and after applying augmentation, DenseNet121 is employed followed by Dropout and AVGPooling techniques as well as LSTM and Global attention module are utilized. Also, RAFT and FlowNet2 are served to extract optical flow. We evaluate our model on Brain4Car Dataset and achieve high performance compared with state-of-the-art studies.

The rest of the paper is organized as follows. In Section 2, the existing literature and state-of-the-art methods are concisely reviewed. The proposed method is described from a technical perspective in Section 3, including the relevant datasets, details of the preprocessing and training procedures, along with the evaluation metrics. In section 4, the results are presented and evaluated against the ones reported in the literature heretofore. Eventually, in Section 5, the conclusions are listed, and suggestions on possible future research directions are put forward.

## 2. Literature review

In this section, several works done in the field of maneuver and driver's intention prediction to improve ADAS systems are introduced. Among the first teams working on autonomous vehicles was the Brain4cars [7] team, which published the first dataset of natural driving. Utilizing information such as tracking the driver's head movement and speed of the car, a model called Autoregressive Input-Output HMM (AIO-HMM) is introduced that can anticipate maneuver within 3.5 seconds before they occur with F1-score rate of over 80% in real-time [7]. The Brain4cars team in another attempt offered a sensory-fusion deep learning architecture based on Recurrent Neural Networks (RNNs) with Long Short-Term Memory (LSTM) units called F-RNN-UL and F-RNN-EL to predict maneuvering on Brain4cars dataset, using both videos (inside and outside the car) and facial landmarks, head pose, car speed, GPS information and lane configuration. Their proposed model learned to fuse multiple sensory streams, and by training it in a sequence-to-sequence prediction manner, it is explicitly learned to anticipate using only a partial temporal context. In order to prevent over-fitting, a novel loss function for anticipation was presented. Their sensory fusion deep learning approach obtained a precision of 84.5% and recall of 77.1%, and anticipated maneuvers in 3.5 seconds (on average) before they happen. By combining the driver's 3D head-pose, the precision and recall improved to 90.5% and 87.4%, respectively [8]. In [9], the proposed prediction system utilized Deep Bidirectional Recurrent Neural Network (DBRNN) consisting of multiple Long-Short Term Memory (LSTM) units and Gated Recurrent Units (GRU) cells that learns to identify the spatial-temporal dependencies in time series data. They used driving environment and the driver (in&out cabin) and evaluated their system for braking, lane change and turn anomaly action prediction on their suggested data and Brain4Cars dataset, resulting in average accuracy of ~80% within 3 second from the braking event. In the paper [10], a novel deep learning architecture and sensory data source such as GPS providing location, car speed, cameras installed inside and outside of the car and other related car sensors presented on Brain4cars dataset are employed. The proposed method introduced a sensory-fusion deep learning framework using a combination of dilated CNN and CNN max-pooling pairs. Convolution and max-pooling pairs were used to learn spatial relationships in video frames and dilated deep convolution structures to record long temporal dependencies, processes, and predicted driver activity. The results of precision and recall were 91.8% and 92.5%, respectively. Zhou et al. [11] presented a Cognitive Fusion Recurrent Neural Networks (CF-RNN) model based on the cognition-driven model and data-driven model. CF-RNN includes two LSTM units that fuse in-cabin and out-cabin data in a cognitive way and by human cognition time process the outputs of this two LSTM units were adjusted, which leading to improvement in F1-score from 88.9% to 92.1% on Brain4Cars dataset. Tonutti et al. [2] applied an LSTM-GRU model for feature extraction and maneuver prediction using driver's head features, eye movement and driving environment along with Domain-Adversarial RNN (DA-RNN) model for domain-adversarial training to achieve domain adaptation. DA-RNN model was evaluated on Brain4cars dataset as the source domain and a proposed new dataset as the target domain. Moreover, to optimize the extraction of domain-independent features, fine-tuning method was employed, which led to the obtained results as follows, for Brain4Cars dataset, precision, recall and F1-score were 92.3%, 90.8% and 91.3%, and for the introduced data were 89.4%, 92.2% and 90.8%, respectively. Unlike previous works that focused on hand-crafted features, in [12] the model anticipated the driver intention directly from videos in an end-to-end method for maneuver. The proposed model consists of three components: a FlowNet2.0 [13] architecture for optical flow extraction to obtain the motion-based representations, a 3D residual network (3D ResNet) for maneuver classification and LSTM unit for handling temporal data of varying length. They fused driver observation data from inside and outside the cabin and fine-tuned the proposed model pre-trained on the large-scale Kinetics dataset, resulting in accuracy rate of 83.12% and F1-score of 81.74% on the Brain4Cars dataset. Inspired by [3], the authors of [14] presented an architecture based on RNN and LSTM, using Brain4Cars dataset that combined both information from inside and outside the car to predict driver's actions, which leading to achieve 92.12%, 87.95%, 95.95% and 86.1% for accuracy, precision, recall and F1-score, respectively.

In [15], Moussaid et al. employed two processing sections to predict the maneuver: In the first section, they focused on feature extraction using a CNN DenseNet121 [16] architecture, followed by obtaining dataframe with 256 attributes combined with the outside features. The second part involves a CNN-LSTM model which is a combination of two standard models CNN and LSTM for maneuvering. To strengthen their model, the anomalies were detected and replaced with more meaningful values along with testing the model by adding noise to the images. The proposed method was able to predict the maneuver with an accuracy of 94.1% in 3.75 seconds before rotation. Gite et al. [4] developed a driver's movement tracking (DMT) algorithm using only inside videos of Brain4cars dataset. To improve the action anticipation performance, a fusion of spatiotemporal data points (STIPs) for driver's movement tracking was introduced and a fast eye gaze algorithm to track eye movements was employed. The features were extracted from STIP and eye gaze were fused and analyzed by a deep recurrent neural network to improve the prediction time. Applying the F-RNN-DMT architecture, the results of 96.21%, 94.11% and 97.56% for accuracy, precision and recall were gained. Zhou et al. [17] first introduced a CF-LSTM model based on cognition-driven method and data-driven method inspired by [8] for feature extraction. This model includes two LSTM units for both internal and external currents of the car, which describes the external features including speed, the lane configuration and internal features, driver head movement, and Driver's face landmark using the CLM or CLNF algorithm [7] is obtained. The authors also introduce an architecture called the Predictive-Bi-LSTM-CRF model and a comprehensive evaluation metric that predicts maneuver with 93.6% for the F1-score metric and accuracy 94.83% accuracy for the proposed metrics on the Brain4cars database. The architecture proposed in [1] utilized two streams of data: outside and inside frames of the car. First, using FlowNet 2.0 [18], optical flow of images are generated from the main out-cabin frames and fed to a ConvLSTM encoder part of model. The features extracted from the optical flow images are then imported into a decoder called Conv layers. In another attempt, using a 3D resnet-50 network, feature extraction was performed from the in-cabin frames, resulting in accuracy of 83.98% and the f1-score of 84.3% on Brain4cars dataset.

The summary of review works is listed in Table 1 in terms of the used dataset, data source, method and evaluation metrics including accuracy, precision, recall and f1-score.

*Table 1. the summary of performance of studies worked on driver maneuver anticipation*

| References | Dataset | Data Source | Method | Accuracy (%) | Precision (%) | Recall (%) | F1-score (%) |
|---|---|---|---|---|---|---|---|
| Rong et al. [1] | Brain4Cars | in<br>out<br>in-&out-side | ConvLSTM auto-encoder & Resnet50 using Flownet2 | 77.40<br>60.87<br>83.98 | - | - | 75.49<br>66.38<br>84.3 |
| Zhou et al. [17] | Brain4Cars | in-&out-side | CF-LSTM & Predictive-Bi-LSTM-CRF | - | 92.4 | 94.7 | 93.6 |
| Gite et al. [4] | Brain4Cars | In-side | F-RNN-DMT | 96.21 | 94.11 | 97.56 | - |
| Moussaid et al.[15] | Brain4Cars | in-&out-side | CNN-LSTM | 94.1 | - | - | - |
| Gite et al. [14] | Brain4Cars | in-&out-side | RNN-LSTM | 92.12 | 87.95 | 95.95 | - |
| Gebert et al. [12] | Brain4Cars | in<br>out<br>in-&out-side | 3D ResNet & LSTM using Flownet2 | 83.1<br>53.2<br>75.5 | - | - | 81.7<br>43.4<br>73.2 |
| Tonutti et al. [2] | Brain4Cars | in-&out-side | DA-RNN & LSTM-GRU | - | 92.3 | 90.8 | 91.3 |
| | Their data | | | | 89.4 | 92.2 | 90.8 |
| Zhou et al. [11] | Brain4Cars | in-&out-side | CF-RNN | - | 92 | 92.3 | 92.1 |
| Rekabdar et al. [10] | Brain4Cars | in-&out-side | Dilated CNN | - | 91.8 | 92.5 | - |
| Olabiyi et al. [9] | Brain4Cars & their data | in-&out-side | DBRNN | Braking: ~80 Lane change: ~80 Turning: ~90 | - | - | - |
| Jain et al. [8] | Brain4Cars | in-&out-side | F-RNN-UL & F-RNN-EL | - | 90.5 | 87.4 | |
| Jain et al. [7] | Brain4Cars | in-&out-side | AIO-HMM | - | 77.4 | 71.2 | ~82 |

## 3. Methods

### 3.1. Dataset

In this study The Brain4Cars [7] dataset is utilized for evaluating the proposed model. Brain4Cars dataset includes two different views videos: (a) driver observation videos and (b) videos of the outside scenes with details of (1088px × 1920px, 25 fps) and (480px × 720px, 30 fps), respectively, that recorded simultaneously [1]. There are five classes of maneuvers in the dataset: *go straight, left lane change, left turn, right lane change, right turn*. According to the Brain4cars dataset, the video covers the behavior before the actual maneuver occurs, i.e., no maneuver is performed during the video. This dataset is collected from 10 drivers with a car equipped with a camera and the videos are annotated in a total of 700 events containing 274 lane changes, 131 turns, and 295 randomly sampled instances of driving straight [1]. In this work, we select and use only 5-second videos. Hence, the number of used videos is 455 with the length of 5-seconds. Which includes 364 videos for training and 91 videos for testing.

### 3.2. Architecture

In this section, the architecture of the model is explained.

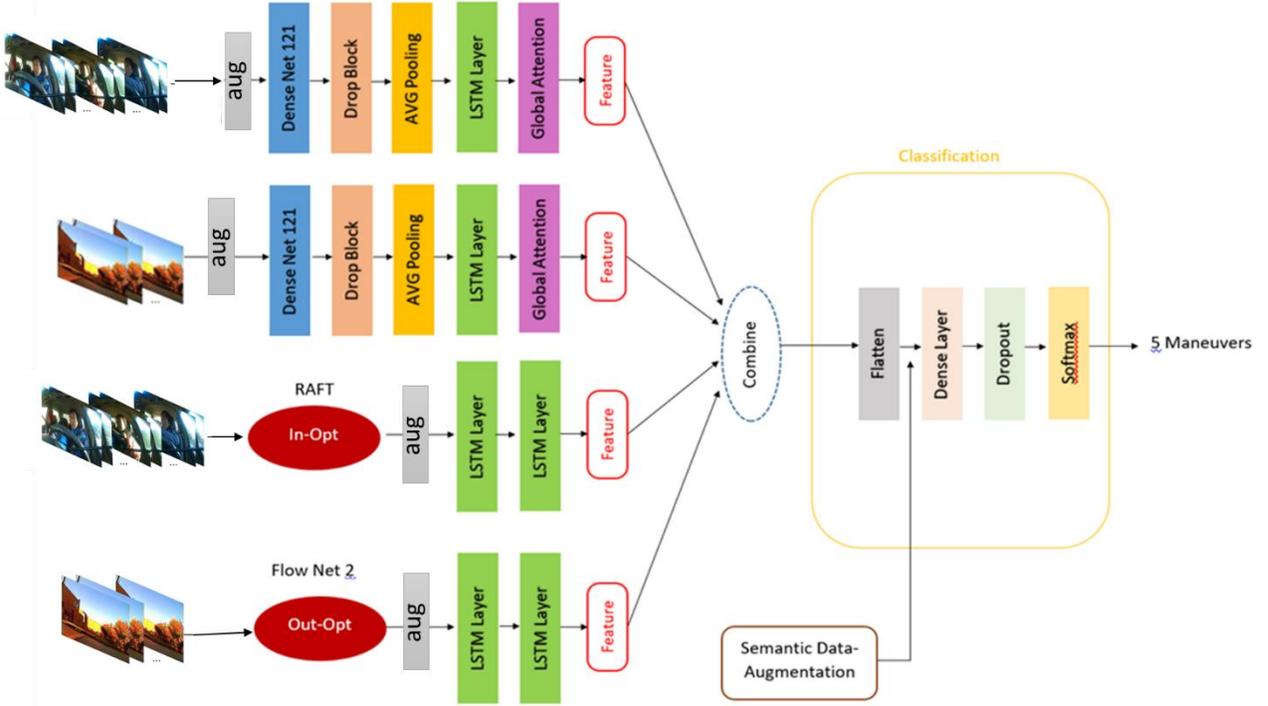

*Figure 1. Schematic illustration of the proposed architecture. In first scenario the branches using outside view are eliminated and in second scenario the branches using inside view are ignored.*

The overview of our framework has been illustrated in Figure 1. The proposed model consists of 4 input sources: main frames inside and outside the cabin, optical flow frames inside and outside the cabin. For each of the 4 input sources, we select the frames with a sample rate of 10, so that each 5-second video consists of 15 frames.

### 3.2.1. Pre-processing and data augmentation

First, the frames of each of the 4 input branches are resized to $128 \times 128$, and then data augmentation is applied to the raw images in first and second branch and optical flow, the output of RAFT and FlowNet2, in third and fourth branch as it depicted in Figure 1. The used augmentations in the proposed method include translation, Flip-Left-to-Right (FlipLR), cutout [19] and a new technique called Augmix [20]. Translation involves moving the image in X or Y direction (or both). In this work, the raw and optical flow image was moved by 4 pixels in both directions. In addition, FlipLR which flip the image horizontally is applied and due to its impact on its label, the label also changed. In other words, when the image showing the driver turning left is augmented by FlipLR, the augmented image shows the driver acting the opposite direction, thereby, the corresponding label is altered to left. Another augmentation technique named cutout is a regularization technique that randomly masks out square regions. Augmix also is a data processing technique, which mixes randomly generated augmentations (autocontrast, equalize, posterize, solorize). More detail and information is described in [20]. Some sample of augmented images have been depicted in Figure 2.

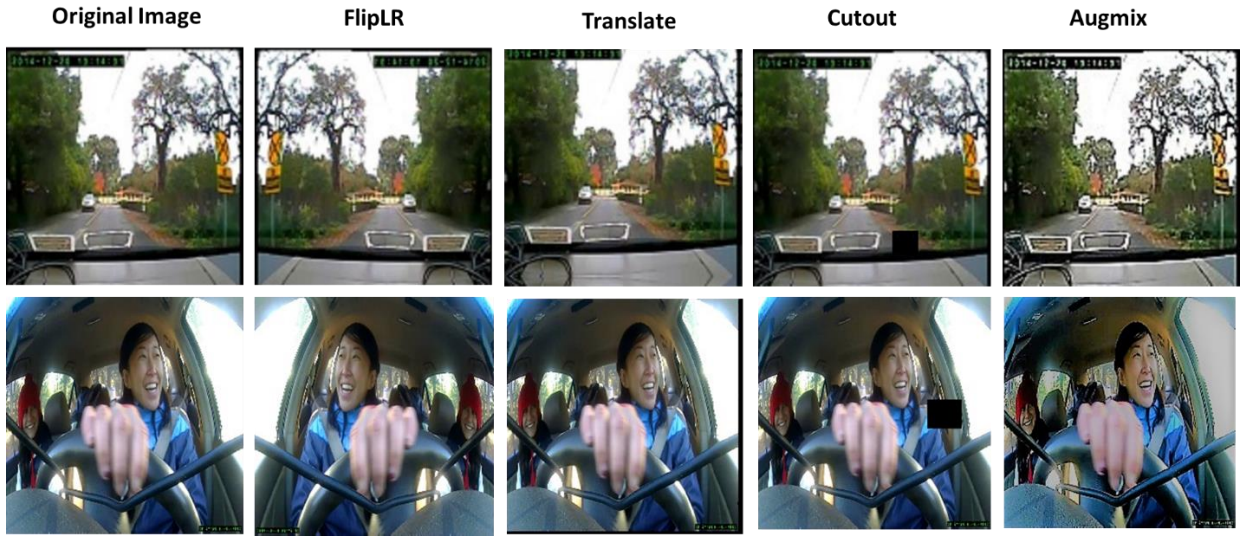

*Figure 2. some samples of applied augmentations (augmix, cutout, translate). Each row illustrates a sample image and the applied augmentation is shown on top of each column.*

### 3.2.2. Feature extraction

Inspired by model in [1] the proposed framework is illustrated in Figure 1. In first two branches, after applying augmentations (Translation, FlipLR, Cutout, Augmix) described in section 3.2.1, the inside view images (in first branch) and the outside view images (in second branch) are normalized and the images inside and outside the cabin are fed to the DenseNet121 model [21] to extract the features. The extracted features from the DenseNet121 module are then passed through a Dropblock layer [22] with the block-size of 5. An AVGPooling layer is then added to the Dropblock output and a LSTM layer with 512 memory units is also followed by a Global Attention layer.

In second two branches, in-cabin optical flow frames produced by Recurrent All-Pairs Field Transforms (RAFT) [23] (in third branch) and out-cabin optical flow frames produced by FlowNet2 [18] (in fourth branch) are augmented by Translation and FlipLR technique described above. Then, the produced data is resized to $128 \times 384$ and fed into two successive LSTM layers with 128 memory cells.

It is noted that in first scenario when only inside view is utilized, the second and fourth branches of model are eliminated. Similarly, when we use outside view in second scenario, the first and third branches are ignored.

### 3.2.3. Classification

In second part of model, the extracted features produced in first part of model are classified. Firstly, all features extracted from all 4 input branches are concatenated and passed through a Flatten layer. Then the implicit semantic data augmentation algorithm (ISDA)[24]which is novel technique for augmentation is applied. In the next step, output of the semantic data augmentation module is passed through a layer of Dense with 512 neurons followed by dropout layer with rate of 0.45. Eventually, the Softmax layer is employed and the probability of the given input is generated.

### 3.2.4. Training

To train our network, the number of epochs is 320 with a batch size of 5, a learning rate of 0.0003 and an Adam [25] optimizer. The network was trained using the categorical cross-entropy loss function and the training process was conducted in Google Colab graphics processing unit.

#### 3.2.4.1. Network evaluation.

To evaluate the performance of the proposed network like [4], we use four standard performance metrics, described in Eq (1) to (4): accuracy, precision, recall, F1-score as well as confusion matrix. The elements to calculate the mentioned metrics are True Positive (TP), True Negative (TN), False Positive (FP), and False Negative (FN), that are defined as follows for the driver action prediction:

- True Positive (TP) = correct action prediction
- False Positive (FP) = incorrect action prediction
- True Negative (TN) = predict action, but the driver does not perform any action (driving straight)
- False Negative (FN) = predict straight driving, but the driver performs the action

$$Precision = TP/(TP + FP) \quad (1)$$

$$Recall = TP/(TP + FN) \quad (2)$$

$$F1 - score = (2 * Precision * Recall)/(Precision + Recall) \quad (3)$$

$$Accuracy = \frac{(TP+TN)}{(TP+TN+FP+FN)} \quad (4)$$

## 4. Results and Discussion

In this section, the performance of proposed method in three different scenarios (in-cabin, out-cabin, in-cabin and out cabin) are presented. The results of early detection capability in seconds are provided in each scenario listed in Table 2 , Table 5 and Table 8 .The action is recognized before time step T which $T \in (-4, -3, -2, -1, 0)$ and the – illustrate the time (in second) before maneuver happens. Furthermore, the obtained results are compared with state-of-the-art studies shown in Table 3, Table 6 and Table 9 in and 5-fold cross validation for all experiment are provided illustrated in Table 4, Table 7 and Table 10. It should be noted that all provided results are in percentage. Finally, the effect of aforementioned augmentations in performance for using in-cabin and out-cabin view as well as confusion matrix for all three scenarios are depicted.

**4.1. In-cabin Action Recognition**

In this section, to predict driver action only the inside images are utilized illustrated in figure 2 in second row. Table 2 is depicted the accuracy, precision, recall and f1-score evaluation metrics. As it can be seen, the accuracy, precision, recall and f1-score increased from 4 seconds before the real action and eventually reach to 89.01%, 89.13%, 89.01% and 88.89%, respectively. In Table 3, our proposed work compared with some presence studies which used only inside view. As it can be seen, our results accuracy of 89.01%, precision of 89.13%, recall of 89.01 and f1-score of 88.89% exceed other works Rong et al. [1] and Gebert et al. [12] , except Gite et al.[4] that obtained 96.21%, 94.11% and 97.56% for accuracy, precision and recall, respectively. In addition, we employed 5-fold cross-validation to guarantee that distribution of training and test data is logical. Table 4 provides the accuracy rate of each step in k cross validation method in which k is equal to 5. In (Original, Translate and Cutout) OTC method the original image, image augmented by translate method and, cutout image is given to the model separately and the max vote is selected as the final result. The mean of all steps as well as standard deviation is listed in the last column. Similar to Table 2 , the accuracy in 0 time is the highest and for -4, -3, -2, -1 and 0 the result fluctuates around 46%, 60%, 75%, 83% and 88 %, respectively. Utilizing the method of OTC is shown better performance in each time step by one or two percent.

*Table 2. The obtained results using only inside view in terms of precision, recall, f1-score and accuracy.*

| Time (s) | Accuracy(%) | Precision(%) | Recall(%) | F1-score(%) |
|---|---|---|---|---|
| 0 | **89.01** | **89.13** | **89.01** | **88.89** |
| -1 | 83.51 | 83.22 | 83.52 | 83.20 |
| -2 | 75.82 | 75.41 | 75.82 | 75.42 |
| -3 | 60.43 | 56.38 | 60.44 | 56.41 |
| -4 | 46.15 | 45.83 | 46.15 | 45.20 |

*Table 3. Comparison of our method with presented state-of-the-art methods using inside view based on accuracy, precision, recall and F1-score.*

| Reference | Accuracy(%) | Precision(%) | Recall(%) | F1-score(%) |
|---|---|---|---|---|
| Rong et al. [1] | 77.40 | - | - | 75.49 |
| Gite et al. [4] | **96.21** | **94.11** | **97.56** | - |
| Gebert et al. [12] | 83.1 | - | - | 81.7 |
| Ours | 89.01 | 89.13 | 89.01 | 88.89 |

*Table 4. Accuracy results of K-fold method using inside view, K=5.*

| Time(s) | method | Fold 1 | Fold 2 | Fold 3 | Fold 4 | Fold 5 | Mean ±std(%) |
|---|---|---|---|---|---|---|---|
| 0 | Our | 87.91 | 87.91 | 89.01 | 87.91 | 89.01 | 88.35 ±.6 |
| 0 | **Our + OTC** | 87.91 | 87.91 | 89.01 | 89.01 | 89.01 | **88.57 ±.6** |
| -1 | Our | 80.21 | 83.51 | 83.51 | 81.31 | 83.51 | 82.41 ±1.5 |
| -1 | **Our + OTC** | 83.51 | 83.51 | 83.51 | 81.31 | 83.51 | **83.07 ±.98** |
| -2 | Our | 75.82 | 74.72 | 73.62 | 74.72 | 75.82 | 74.94 ±.92 |
| -2 | **Our + OTC** | 75.82 | 74.72 | 74.72 | 75.82 | 75.82 | **75.38 ±.60** |
| -3 | Our | 59.34 | 60.43 | 58.24 | 59.34 | 60.43 | 59.56 ±.91 |
| -3 | **Our + OTC** | 60.43 | 61.53 | 60.43 | 59.34 | 60.43 | **60.43 ±.77** |
| -4 | Our | 43.95 | 45.05 | 45.05 | 43.95 | 46.15 | 44.83 ±.92 |
| -4 | **Our + OTC** | 48.35 | 45.05 | 46.15 | 46.15 | 46.15 | **46.37 ±1.2** |

## 4.2. Out-cabin Action Recognition

In second scenario, the model trained with only outside view images. Following the same pattern in Table 2, the performance upsurges when approaching to 0 second represented in Table 5. In 4 seconds before real action, the accuracy precision, recall and f1-score were 42.86% , 44.18% , 42.86% and 42.84%  and soar in each time step and reach to 82.41%, 82.28%, 82.42% and 82.24% in 0 second. In Table 6, we compare our model with state-of-the-art models which used only outside view images, observing surpassing other works. The accuracy, precision, recall and f1-score of our model are 82.41%, 82.28%, 82.42% and 82.24%, while [1] achieved accuracy of 60.87% and f1-score of 66.38%, and [12] obtained accuracy of 53.2% and f1-score 43.4%. Similarly, to calculate accuracy k cross validation method is also applied which is shown in Table 7. The parameter K is equal to 5 and OTC method described in 4.1 section is utilized. In 0 second, using OTC, the accuracy of 82.19% is obtained which is close to what achieved in Table 5, 82.24%.

Table 5. The obtained results using only outside view in terms of precision, recall, f1-score and accuracy.

| Time(s) | Accuracy(%) | Precision(%) | Recall(%) | F1-score(%) |
|---|---|---|---|---|
| **0** | **82.41** | **82.28** | **82.42** | **82.24** |
| -1 | 76.92 | 76.65 | 76.92 | 76.64 |
| -2 | 67.03 | 64.20 | 67.03 | 64.28 |
| -3 | 56.04 | 51.04 | 56.04 | 51.50 |
| -4 | 42.86 | 44.18 | 42.86 | 42.84 |

Table 6. Comparison of our method with presented state-of-the-art methods using outside view based on accuracy, precision, recall and F1-score.

| Reference | Accuracy(%) | Precision(%) | Recall(%) | F1-score(%) |
|---|---|---|---|---|
| Rong et al. [1] | 60.87 | - | - | 66.38 |
| Gebert et al. [12] | 53.2 | - | - | 43.4 |
| **Ours** | **82.41** | **82.28** | **82.42** | **82.24** |

Table 7. Accuracy results of K-fold method using only outside view, K=5.

| Time(s) | method | Fold 1 | Fold 2 | Fold 3 | Fold 4 | Fold 5 | Mean ±std(%) |
|---|---|---|---|---|---|---|---|
| 0 | Our | 80.21 | 81.31 | 81.31 | 82.41 | 82.41 | 81.53±.92 |
| 0 | **Our + OTC** | 82.41 | 82.41 | 81.31 | 82.41 | 82.41 | **82.19±.49** |
| -1 | Our | 75.82 | 76.92 | 75.82 | 75.82 | 76.92 | 76.26±.60 |
| -1 | **Our + OTC** | 75.82 | 76.92 | 76.92 | 75.82 | 76.92 | **76.48±.60** |
| -2 | Our | 65.93 | 65.93 | 64.83 | 65.93 | 67.03 | 65.93±.77 |
| -2 | **Our + OTC** | 67.03 | 65.93 | 67.03 | 68.13 | 67.03 | **67.03±.77** |
| -3 | Our | 54.94 | 53.84 | 56.04 | 56.04 | 56.04 | 55.38±.98 |
| -3 | **Our + OTC** | 56.04 | 56.04 | 56.04 | 57.14 | 56.04 | **56.26±.49** |
| -4 | Our | 41.75 | 41.75 | 40.65 | 41.75 | 42.86 | 41.75±.78 |
| -4 | **Our + OTC** | 41.75 | 42.86 | 42.86 | 43.95 | 42.86 | **42.86±.77** |

### 4.3. In-cabin and Out-cabin Action Recognition

In this scenario, we utilized in-cabin and out-cabin images to train our model, resulting in best performance comparing two previous scenarios (in-cabin and out-cabin). Table 8 shows the accuracy of 98.90%, precision of 98.96%, recall of 98.90% and f1-score of 98.88% of this scenario in real time action. In Table 9, we compare our result with other works and it can be seen that most of present work use both inside and outside views to enhance the performance. Our results were 98.90%, 98.96%, 98.90% and 98.88% for accuracy, precision, recall and f1-score, outperforming other papers listed in Table 9. Table 10 also provides k cross validation method similar to previous scenarios. In second 0, we achieved accuracy of 98.46% which is close to obtained result in Table 8. As aforementioned, OTC is served to enhance the performance resut.

*Table 8. The obtained results using inside and outside view in terms of precision, recall, f1-score and accuracy.*

| Time(s) | Accuracy (%) | Precision (%) | Recall (%) | F1-score (%) |
|---|---|---|---|---|
| **0** | **98.90** | **98.96** | **98.90** | **98.88** |
| -1 | 93.40 | 93.58 | 93.41 | 93.39 |
| -2 | 84.61 | 84.67 | 84.62 | 84.51 |
| -3 | 71.42 | 70.24 | 71.43 | 70.22 |
| -4 | 57.14 | 52.04 | 57.14 | 52.20 |

*Table 9. Comparison of our method with presented state-of-the-art methods using both inside and outside view based on accuracy, precision, recall and F1-score.*

| Reference | Accuracy(%) | Precision(%) | Recall(%) | F1-score(%) |
|---|---|---|---|---|
| Rong et al. [1] | 83.98 | - | - | 84.3 |
| Zhou et al. [17] | - | 92.4 | 94.7 | 93.6 |
| Moussaid et al. [15] | 94.1 | - | - | - |
| Gite et al. [14] | 92.12 | 87.95 | 95.95 | - |
| Gebert et al. [12] | 75.5 | - | - | 73.2 |
| Tonutti et al. [2] | - | 92.3 | 90.8 | 91.3 |
| Zhou et al. [11] | - | 92 | 92.3 | 92.1 |
| Rekabdar et al. [10] | - | 91.8 | 92.5 | - |
| Jain et al. [8] | - | 90.5 | 87.4 | |
| Jain et al. [7] | - | 77.4 | 71.2 | Over 80 (82) |
| **ours** | **98.90** | **98.96** | **98.90** | **98.88** |

Table 10. Accuracy results of K-fold method using both inside and outside view, K=5.

| Time(s) | method | Fold 1 | Fold 2 | Fold 3 | Fold 4 | Fold 5 | Mean ±std(%) |
|---|---|---|---|---|---|---|---|
| 0 | Our | 96.70 | 97.80 | 97.80 | 96.70 | 98.90 | 97.58±0.92 |
| 0 | **Our + OTC** | 97.80 | 97.80 | 98.90 | 98.90 | 98.90 | **98.46±0.60** |
| -1 | Our | 91.20 | 93.40 | 92.3 | 91.20 | 93.40 | 92.3±1.1 |
| -1 | **Our + OTC** | 92.30 | 93.40 | 92.30 | 92.30 | 94.50 | **92.96±0.98** |
| -2 | Our | 82.41 | 83.51 | 82.41 | 83.51 | 84.61 | 83.29±0.92 |
| -2 | **Our + OTC** | 83.51 | 82.41 | 82.41 | 84.61 | 85.71 | **83.73±1.43** |
| -3 | Our | 68.18 | 69.23 | 70.32 | 67.03 | 71.42 | 69.24±1.72 |
| -3 | **Our + OTC** | 69.23 | 71.42 | 70.32 | 67.03 | 71.42 | **69.88±1.83** |
| -4 | Our | 49.45 | 54.94 | 52.74 | 50.54 | 57.14 | 52.96±3.14 |
| -4 | **Our + OTC** | 50.54 | 54.94 | 52.74 | 51.64 | 58.24 | **53.62±3.05** |

### 4.4. Effect of Augmentation on Performance

As described in 3.2.1 section, different augmentation methods (FlipLR, Translate, Cutout and Augmix) are utilized in this study. Figure 3 illustrates the effect of augmentation on performance in which top left, top right, left down, right down images illustrate accuracy, precision, recall and f1-score, respectively. Plot A shows the model without employing any augmentation and in plot B, FlipLR is added to A. Plot C surpasses the previous ones (A and B) which Cutout is added to B. Similarly, Augmix augmentation is added to C named D. Plot E which is used all mentioned augmentations (FlipLR, Cutout, Augmix, Translate) excels all others explained methods.

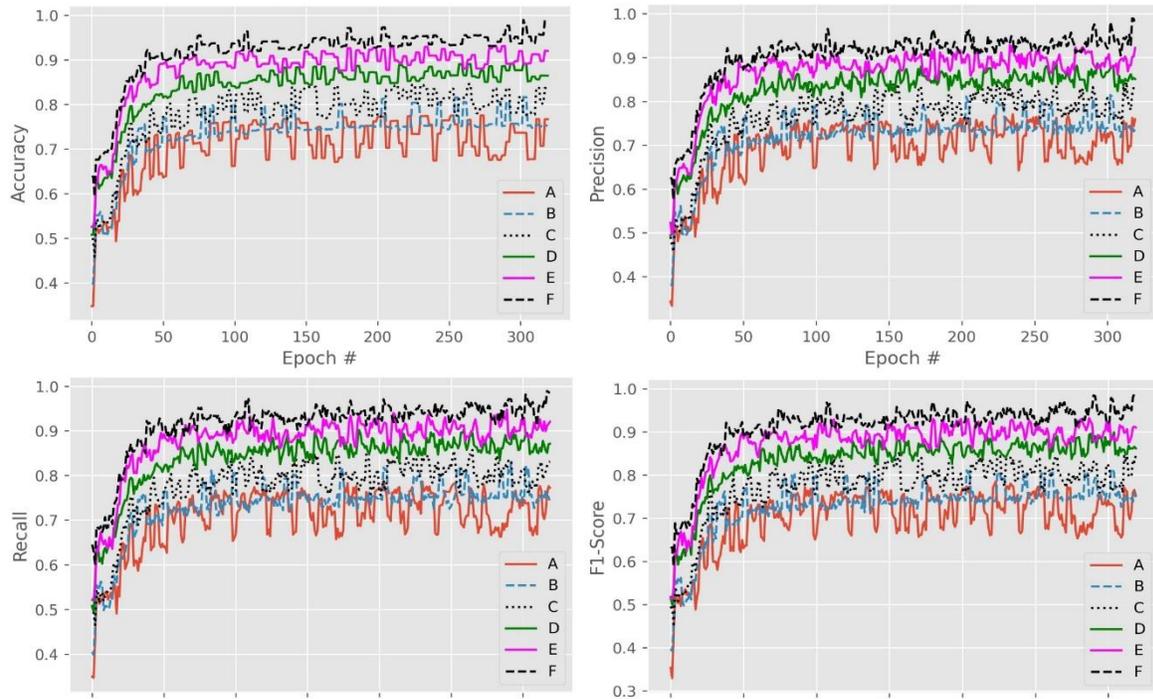

Figure 3. The effect of using augmentation on accuracy, precision, recall and f1-score. A = base, B = base+FlipLR, C = B+cutout, D = C + Augmix , E = D+ smooth + translate ,

## 4.5. Confusion Matrix

Moreover, to represent the performance of proposed architecture, the confusion matrixes are depicted in Figure 4, where the numbers in diagonal stand for the count of correctly recognized samples from corresponding classes. It is observable that the model error in confusion matrix (c) showing the model trained with both in-cabin and out-cabin views are less than other two matrixes (a) and (b) representing the model trained with inside and outside view, respectively. Also, they indicate the fact that the model's error is not biased toward any of the specific classes, but is rather distributed to all of the classes.

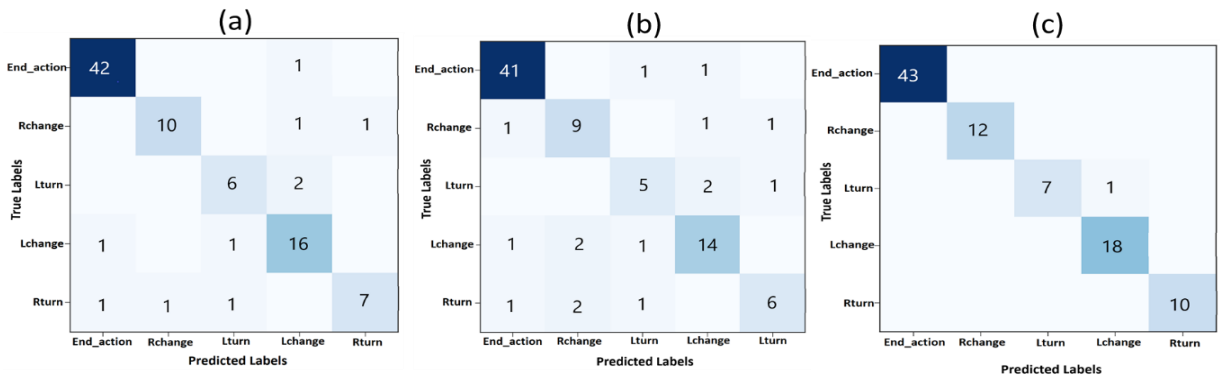

Figure 4. Fusion matrix for (a) inside view (b) outside view (c) inside and outside view

## 5. Conclusion

In this study, a new deep neural network is proposed to anticipate the driver maneuver intention. We examined the model in three different scenarios: using only inside view, only outside view and both inside and outside view. In our proposed method, we used DenseNet121, LSTM and global attention module to extract features. Also, RAFT and FlowNet2 are employed to extract optical flow. In first scenario, in order to avoid overfitting and enhancing the performance of proposed framework, different augmentation methods FlipLR, Translate, Cutout and Augmix are served. Utilizing accuracy, precision, recall and f1-score as evaluation metrics, our proposed framework outperform other present state-of-the-art work when outside and inside-outside view is taking into consideration.

For our future work, to improve performance, we aim to use Swin transformer [26] as an encoder to extract image features in fast speed which is essential component in recognizing diver maneuver.